\documentclass{article}
\usepackage{ijcai23}

\usepackage{times}
\usepackage{soul}
\usepackage{url}
\usepackage[hidelinks]{hyperref}
\usepackage[utf8]{inputenc}
\usepackage[small]{caption}
\usepackage{graphicx}
\usepackage{amsmath}
\usepackage{amsthm}
\usepackage{booktabs}
\usepackage{algorithm}
\usepackage{algorithmic}
\usepackage[switch]{lineno}
\usepackage{amsmath}
\usepackage{amssymb}
\usepackage{mathtools}
\usepackage{amsthm}
\usepackage{amsmath}

\newcommand{\indep}{\perp \!\!\! \perp}

\theoremstyle{definition}

\newtheorem{assumption}{Assumption}
\theoremstyle{remark}

\title{Causal Effect Estimation: Recent Advances, Challenges, and Opportunities}

\author{
Zhixuan Chu$^1$
\and
Jianmin Huang$^1$\and
Ruopeng Li$^1$\and
Wei Chu$^1$\and
Sheng Li$^2$
\affiliations
$^1$Ant Group\\
$^2$University of Virginia
\emails
chuzhixuan.czx@alibaba-inc.com,
\{caiqiong, ruopeng.lrp, weichu.cw\}@antgroup.com,
shengli@virginia.edu
}

\begin{document}

\maketitle

\begin{abstract}
Causal inference has numerous real-world applications in many domains, such as health care, marketing, political science, and online advertising. Treatment effect estimation, a fundamental problem in causal inference, has been extensively studied in statistics for decades. However, traditional treatment effect estimation methods may not well handle large-scale and high-dimensional heterogeneous data. In recent years, an emerging research direction has attracted increasing attention in the broad artificial intelligence field, which combines the advantages of traditional treatment effect estimation approaches (e.g., propensity score, matching, and reweighing) and advanced machine learning approaches (e.g., representation learning, adversarial learning, and graph neural networks). Although the advanced machine learning approaches have shown extraordinary performance in treatment effect estimation, it also comes with a lot of new topics and new research questions. In view of the latest research efforts in the causal inference field, we provide a comprehensive discussion of challenges and opportunities for the three core components of the treatment effect estimation task, i.e., treatment, covariates, and outcome. In addition, we showcase the promising research directions of this topic from multiple perspectives.
\end{abstract}

\section{Introduction}

Causality is naturally and widely used in various disciplines of science to discover causal relationships among variables and estimate causal effects of interest. The most effective way of inferring causality is to conduct a randomized controlled trial, randomly assigning participants to a treatment group or a control group. As the randomized study is conducted, the only expected difference between the control and treatment groups is the outcome variable being studied. However, in reality, randomized controlled trials are always time-consuming and expensive. In addition, ethical issues also need to be considered in most randomized controlled trials, which essentially limits its applications. 
Therefore, observational data provide a tempting shortcut instead of randomized controlled trials. Observational data are obtained by the researcher simply observing the subjects without interference. That means the researchers have no control over treatments and subjects and study the subjects by simply analyzing the recorded data. For causal inference, we want to answer questions like ``Would this patient have different results if she received a different medication?'' Answering such counterfactual questions is challenging due to two reasons. First, we only observe the factual outcome and never the counterfactual outcomes that would potentially have happened if the subjects were assigned different treatments. The second one is that treatments are typically not assigned randomly in observational data, which may lead to the treated population differing significantly from the general population, i.e., the well-known selection bias problem.

In recent years, the magnificent bloom of the machine learning area has enhanced the development of causal inference approaches. Powerful machine learning methods, such as decision trees, representation learning, deep neural networks, adversarial learning, and so on, have been applied to estimate the potential outcomes more accurately. In addition to ameliorating the outcome estimation model, machine learning methods also provide a new aspect of handling different types of treatments, leveraging various types of covariates, and mitigating selection bias in different forms. Benefiting from the deep bonding between causal inference and machine learning methods, the treatment effect estimation task has greatly progressed. However, in view of the latest research efforts in the causal inference field, we conclude three major challenges from the core components of the treatment effect estimation task, i.e., treatment, covariates, and outcome:

\begin{itemize}
  \item \textbf{[Treatment]:} How could we deal with different types of treatment, such as (1) binary, (2) multiple, (3) continuous scalar treatments, (4) interrelated sequential treatments, and (5) structured treatments (e.g., graphs, images, texts)?
  \item \textbf{[Covariate]:} How could we handle the different types of covariates, such as confounders (observed and hidden), adjustment, instrumental, and spurious variables by representation disentanglement, feature selection, and so on?
  \item \textbf{[Outcome]:} When estimating the factual and counterfactual outcomes, how could we overcome the selection bias among different treatment groups (for example, distribution invariance, domain adaptation, local similarity, domain overlap, mutual information, and so on)? 
\end{itemize}

\begin{figure}
  \centering
  \includegraphics[width=1\columnwidth]{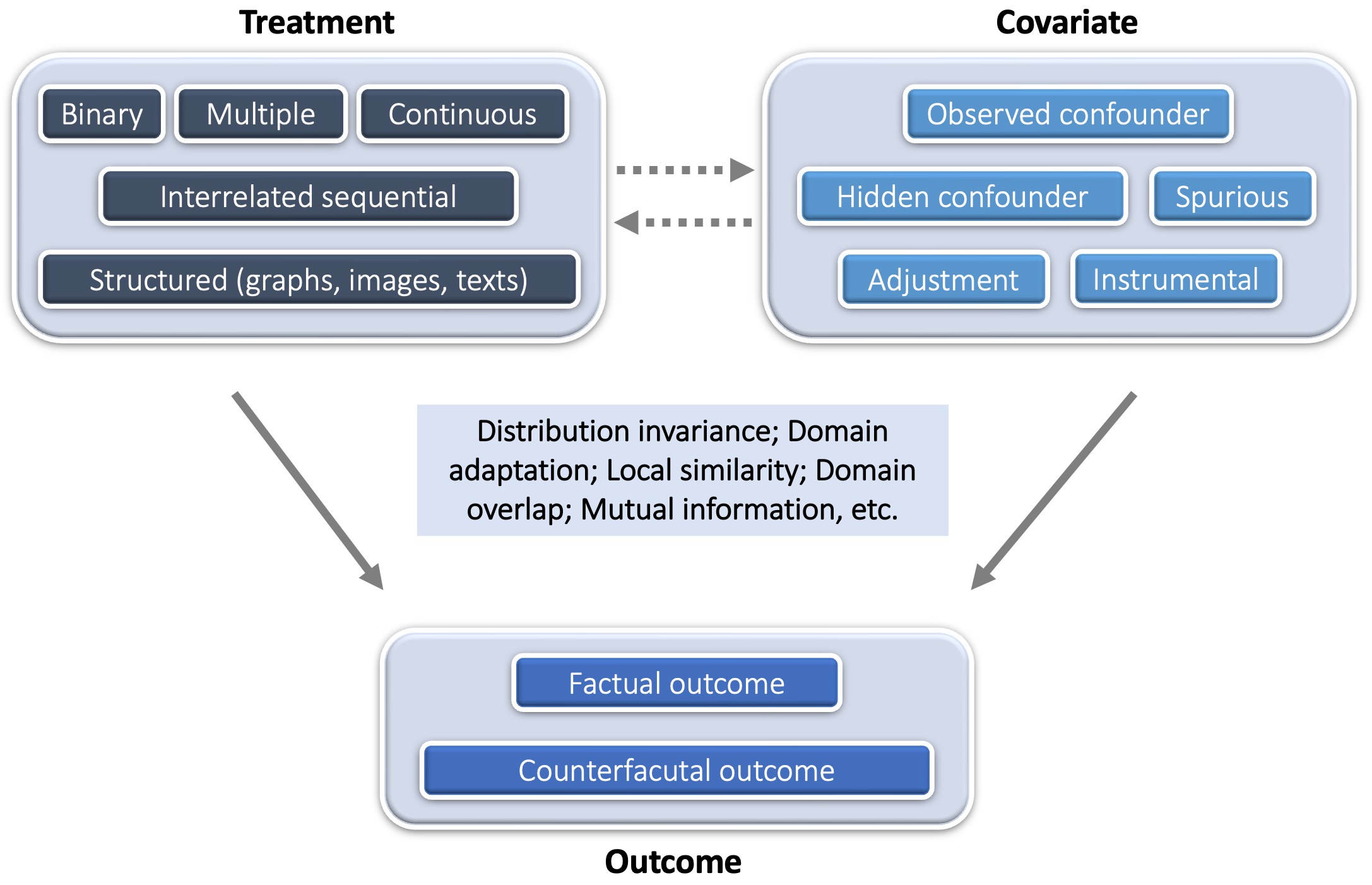}
  \caption{\label{fig:framework} Three major challenges from the core components of the treatment effect estimation task, including treatment, covariates, and outcome.
  }
\end{figure}

As shown in Figure \ref{fig:framework}, different from the previous surveys based on the taxonomy of the methodologies for treatment effect estimation, to the best of our knowledge, this is the first paper that provides a comprehensive survey of challenges abreast of the current academic frontier of treatment effect estimation tasks. In our work, we first consolidate and summarize the previous survey papers about causal inference in Section 2. Then, we formally define the problem of treatment effect estimation in Section 3. Further, from three points of view, i.e., treatment, covariate, and outcome, we describe the research challenges, present the recent advances on these topics, and discuss research opportunities in Section 4. Finally, we conclude this survey paper in Section 5.

\section{Previous Work}

There exist several surveys that discuss a particular category of the causal effect estimation methods, such as the survey of matching-based methods~\cite{stuart2010matching_review}, the survey of tree-based and ensemble-based methods~\cite{athey2015machine}, and the review of dynamic treatment regimes~\cite{chakraborty2014dynamic}. For the structural causal model, it is suggested to refer to the survey~\cite{pearl2009causal} or the book~\cite{pearl2000causality}. There is also a survey about learning causality from observational data~\cite{guo2018survey} whose content ranges from inferring the causal graph from observational data, structural causal model, potential outcome framework, and their connection to machine learning. \cite{cui2020causal} introduce both traditional and state-of-the-art representation learning algorithms for treatment effect estimation and showcase promising applications of these methods in different application domains. A recent survey paper \cite{yao2021survey} mainly focuses on the theoretical background of the potential outcome framework, the representative methods across the statistic domain and machine learning domain, and how this framework and the machine learning area enhance each other. \cite{zeng2022survey} provide a review of the past work on causal inference, focusing mainly on potential outcomes framework and causal graphical models. \cite{feder2022causal} consolidates research on causality in NLP and provides unified definitions, benchmark datasets, and clear articulations of the challenges and opportunities in applying causal inference to the textual domain. \cite{ma2022learning} envision the challenges of causal effect estimation with graphs and then summarize representative approaches of causal effect estimation with graphs in recent years. In addition, the existing causal inference methods mainly focus on source-specific and stationary observational data. Such learning strategies assume that all observational data are already available during the training phase and from only one source. However, in the era of big data, we face new challenges in causal inference with observational data, i.e., the extensibility for incrementally available observational data, the adaptability for extra domain adaptation problems except for the imbalance between treatment and control groups, and the accessibility for an enormous amount of data. The position paper \cite{chu2023continual} formally defines this kind of problem about continual treatment effect estimation, describes its research challenges, and then presents possible solutions to this problem. 

Different from the aforementioned surveys based on traditional methodology-based taxonomy for causal inference, our paper might be the first attempt to provide a comprehensive survey of challenges aiming at the core components of the treatment effect estimation task, i.e., treatments, covariates, and outcomes, abreast of the current academic frontier of treatment effect estimation tasks.

\section{Problem Definition}

Suppose that the observational data contain $n$ units, and each unit received one of the treatments. Let $X$ denote all observed variables, $Y$ denote the outcomes, and $T$ be the treatment in the observational data.  

Let $t_i$ denote the treatment assignment for unit $i$; $i=1,...,n$. Under the basic binary treatment case, $t_i = 1$ is for the treatment group and $t_i=0$ for the control group. The outcome for unit $i$ is denoted by $y_{t}^i$ when treatment $t$ is applied to unit $i$. For observational data, only one of the potential outcomes is observed. The observed outcome is called the factual outcome, and the remaining unobserved potential outcomes are called counterfactual outcomes. 

To estimate the treatment effect from observational data, potential outcome framework ~\cite{splawa1990application,rubin1974estimating} is one of the most popular methodologies, which aims to estimate such potential outcomes and then calculate the treatment effect. The individual treatment effect (ITE) for unit $i$ is the difference between the potential treated and control outcomes and is defined as: 
\begin{equation}
  \text{ITE}_i = y_1^i - y_0^i.   
\end{equation}

The average treatment effect (ATE) is the difference between the mean potential treated and control outcomes, which is defined as: 
\begin{equation}
\text{ATE}=\frac{1}{n}\sum_{i=1}^{n}(y_1^i - y_0^i). 
\end{equation}

The success of the potential outcome framework is based on the following assumptions~\cite{imbens2015causal}, which ensure that the treatment effect can be identified. 

\begin{assumption}
\textit{Stable Unit Treatment Value Assumption (SUTVA)}: The potential outcomes for any unit do not vary with the treatments assigned to other units, and, for each unit, there are no different forms or versions of each treatment level, which lead to different potential outcomes. 
\end{assumption}

\begin{assumption}
\textit{Consistency}: The potential outcome of treatment $t$ is equal to the observed outcome if the actual treatment received is $t$. 
\end{assumption}

\begin{assumption}
\textit{Positivity}: For any value of $x$, treatment assignment is not deterministic, i.e.,$P(T = t | X = x) > 0$, for all $t$ and $x$. 
\end{assumption}

\begin{assumption}
\textit{Ignorability}: Given covariates, treatment assignment is independent of the potential outcomes, i.e., $(y_1, y_0) \indep t | x$. 
\end{assumption}

These assumptions are fundamental to treatment effect estimation. However, in practice, for different types of treatments (multiple, continuous, interrelated sequential, or structured treatments), the existence of hidden confounders, and various methodologies, these assumptions cannot always hold. They shall be relatively relaxed or modified according to practical considerations.

\section{Research Challenges and Opportunities}

In this section, we detail the new challenges regarding treatments, covariates, and outcomes, present the latest research methodologies based on machine learning for these challenges, and also discuss potential research opportunities.

\subsection{Treatment}

We first elaborate on the difficulties when facing different types of treatment, such as binary, multiple, continuous scalar treatments, interrelated sequential treatments, and structured treatments (e.g., graphs, images, texts). According to the characteristics of various treatment types, we will present them in two parts: (1) binary, multiple, continuous, and interrelated sequential treatments; (2) structured treatments.

\subsubsection{4.1.1 Binary, Multiple, Continuous, and Sequential Treatments} 

\begin{figure}
  \centering
  \includegraphics[width=0.9\columnwidth]{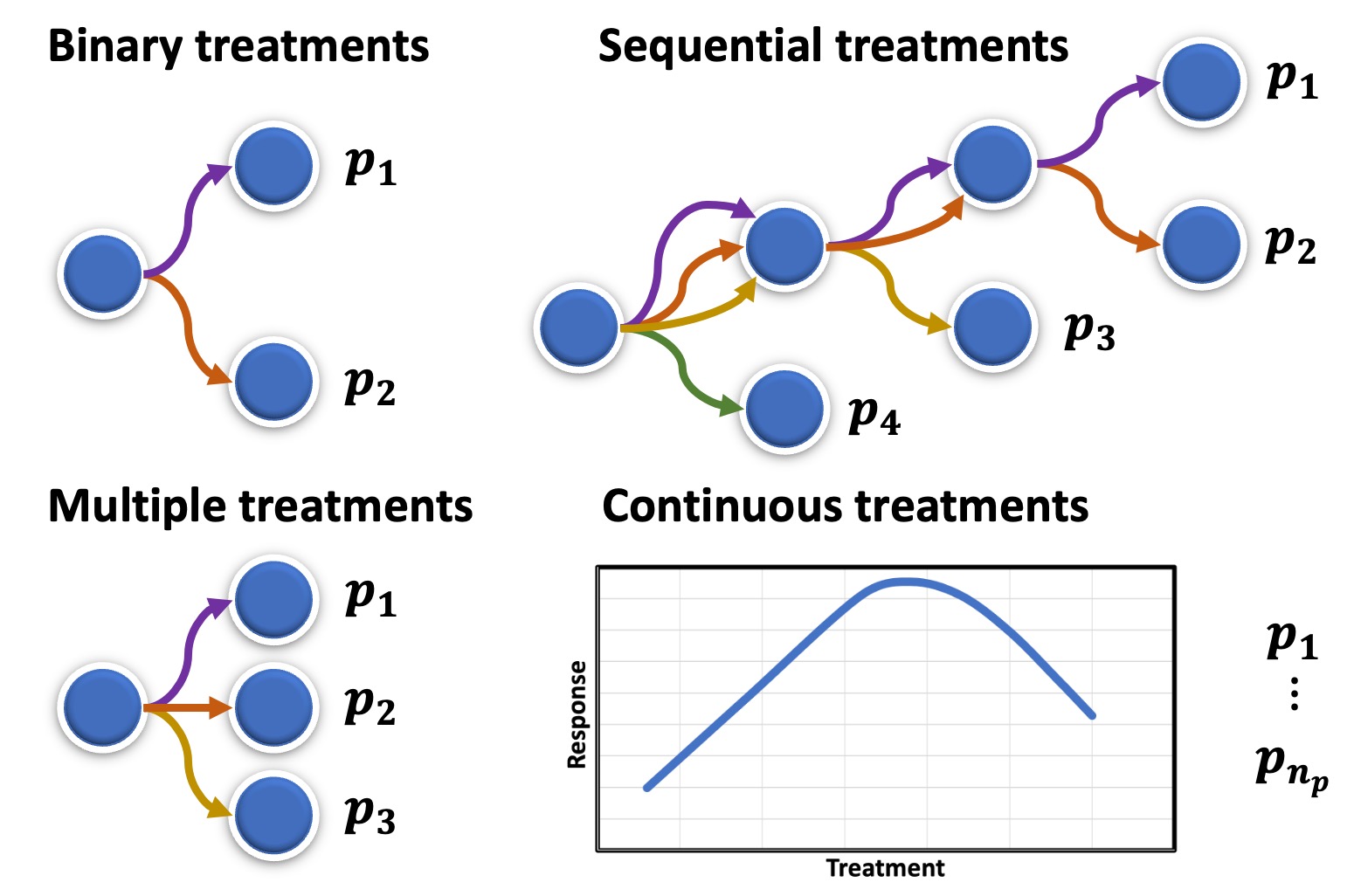}
  
  \caption{\label{fig:treatments} The illustrations of binary, multiple, continuous, and sequential treatments.
  }
\end{figure}

As shown in Figure \ref{fig:treatments}, for the binary, multiple, continuous, and sequential treatment scenarios, we provide a unifying terminology that will enable researchers to coalesce and compare existing methods. Suppose that the observational data contain $n$ units and that each unit goes through one potential path, including several treatment stages. In each potential path, the unit $i$ can sequentially choose one of the two or multiple treatments $T$ at each stage $S$, and finally, the corresponding outcome $Y$ could be observed at the end of the path. Let $\{t_{s}^{i}; t_{s}=1,..., n_{t_{s}}, \ i=1,...,n, \ \text{and}\ s=1,...,n_s\}$ denote the treatment assignment for unit $i$ at stage $s$. There are in total $n_s$ treatment stages and $n_{t_s}$ treatment assignments at stage $s$. Due to the existence of different treatment assignments at each treatment stage, for the whole population, we can observe several potential paths $\{p; p=1,...,n_p\}$. However, each unit can only go through one potential path, including a sequence of stages. Therefore, only one of the potential outcomes is observed at the end of the path according to the actual treatment assignments. This observed outcome is called the factual outcome, and the remaining unobserved potential outcomes are called counterfactual outcomes. The factual outcome for unit $i$ along the actual treatment stages is denoted by $y^{i}_{F}$, and the counterfactual outcome is denoted by $y^{i}_{CF}$. Let $X \in \mathbb{R}^d$ denote $d$ observed variables of a unit. The observational data can be denoted as $\{\{x^{i},\ t_{s}^{i},\ y^{i}_{F}\}_{s=1}^{n_s}\}_{i=1}^n$. For simplicity, the unit superscript $i$ will be omitted unless explicitly needed.

\noindent\textbf{Binary Treatments.} If ${n_s}=1$ and $n_{t_1}=2$, there is only one treatment stage with two treatment choices. A unit only needs to choose once, between the two treatments. This setting is exactly the conventional binary treatment effect estimation task. One practical example of this conventional task is to evaluate the treatment effects of two different medications for one disease. By exploiting the observational data, including the treatment and control groups, we can only get one factual outcome for each patient. Thus the core task is to predict what would have happened if a patient had taken the other medication. This conventional task has been extensively studied in the literature, such as TARNet~\cite{johansson2016learning}, CFR \cite{shalit2017estimating}, BNR-NNM~\cite{li2017matching}, CEVAE~\cite{louizos2017causal}, SITE~\cite{yao2018representation}, GANITE~\cite{yoon2018ganite}, and Dragonnet \cite{shi2019adapting}. 

\noindent\textbf{Multiple Treatments.} If ${n_s}=1$ and $n_{t_1}>2$, there is only one treatment stage with multiple treatments. This is the conventional multiple treatment effect estimation task. Usually, the binary treatment models can be effortlessly extended to multiple treatment models \cite{lopez2017estimation}, such as propensity score estimation using generalized boosted models~\cite{mccaffrey2013tutorial}, the counterfactual inference based on the idea of augmenting samples within a minibatch with their propensity-matched nearest neighbors \cite{schwab2018perfect}, multi-task adversarial learning \cite{chu2022multi}, BART\cite{hu2020estimation}, and a deep generative model with task embedding \cite{saini2019multiple}.

\noindent\textbf{Continuous Treatments.} If ${n_s}\geq 1$ and $t_s$ is continuous, this is the continuous treatment effect estimation task. Continuous treatments arise in many fields, including healthcare, public policy, and economics. With the widespread accumulation of observational data, estimating the average dose-response function while correcting for confounders has become a critical problem. Due to the infinite counterfactuals for continuous treatments, adjusting for selection bias is significantly more complex than for binary or multiple treatments. Thus, unlike the multiple treatments, standard methods for adjusting for selection bias for discrete treatments cannot be easily extended to handle bias in the continuous setting.

The DRNet \cite{schwab2020learning} consists of a three-level architecture with shared layers for all treatments, multi-task layers for each treatment, and additional multi-task layers for dosage sub-intervals. Specifically, for each treatment, the dosage interval is subdivided into several equally sized sub-intervals, and a multi-task head is added for each sub-interval. DRNets do not determine these intervals dynamically and thus much of this flexibility is lost. SCIGAN \cite{bica2020estimating} is flexible and capable of simultaneously estimating counterfactual outcomes for several different continuous interventions. The key idea is to use a modified GAN model to generate counterfactual outcomes. VCNet \cite{nie2021vcnet} proposes a novel varying coefficient neural network that improves model expressiveness while preserving the continuity of the estimated average dose-response function. Second, to improve finite sample performance, we generalize targeted regularization to obtain a doubly robust estimator of the dose-response curve. CausalEGM \cite{liu2022causalegm} is an encoding generative model that can be applied in binary and continuous treatment settings. The CausalEGM model consists of a bidirectional transformation module and two feed-forward neural networks. The bidirectional transformation module composed of two generative adversarial networks (GANs) is used to project the covariates to a low-dimensional space and decouple the dependencies.

\noindent\textbf{Sequential Treatments.}
\label{new_task}
If ${n_s}>1$ and $n_{t_s}\geq2$, there are several treatment stages, with two or multiple treatments at each stage. Each unit goes through one path and needs to make $n_s$ treatment decisions. At the end of the path, we can only observe one outcome along the actual path. 
\begin{figure}
  \centering
  \includegraphics[width=0.85\columnwidth]{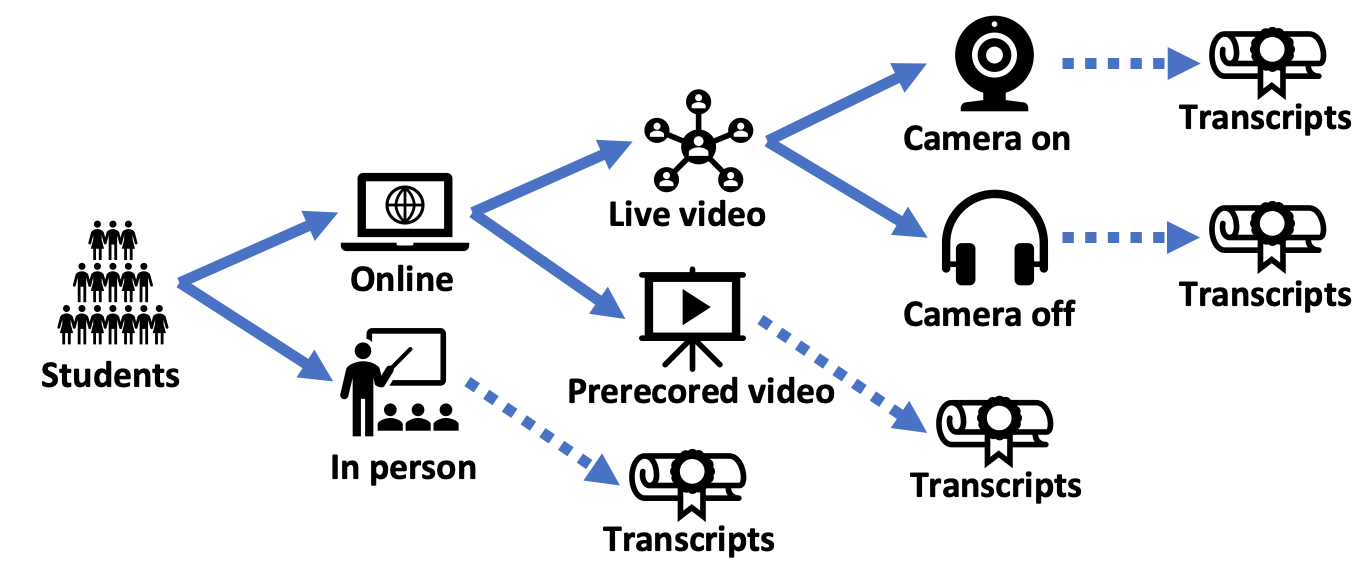}
  
  \caption{\label{fig:example_covid} The instruction modes example. The solid line represents each student's potential choice at each stage, and the dotted line refers to the final potential outcome along the corresponding path.
  }
\end{figure}

For example, during the COVID-19 pandemic that began in late 2019 and continues today, the instruction mode in universities has experienced substantial changes. The COVID-19 pandemic has forced most educational institutes worldwide to resort to an ``online + in person'' mode of education delivery. In some universities, the students can choose online remote learning or in-person learning with masks and social distancing. The course instructors can provide live video-based sessions for the students and/or upload their recordings to the online learning platforms for them to watch. Further, in live video-based learning, the students can choose to turn the camera on or off. Therefore, each student will follow one sequential behavior path ``in person or online learning  $\rightarrow$ prerecorded video-based or live video-based learning $\rightarrow$ camera on or off'', as illustrated in Figure~\ref{fig:example_covid}. Different instruction modes influence students' social, emotional, and mental well-being and academic achievement. Each student makes their own choices at each stage, so various potential paths exist. Intuitively, potential paths are a series of possible choices of treatments for one unit. Each unit only can actually go through one path, which is captured in the observational data. However, at each intervention stage, the unit can choose one of the two or multiple interventions, leading to multiple potential paths, including one factual path and several counterfactual ones. In the causal effect estimation task, we need to estimate the potential outcomes along all potential paths. 

In the circumstances, the selection bias will accumulate and accumulate over multiple stages, making the estimation of counterfactual outcomes more challenging. To the best of our knowledge, existing treatment effect estimation methods cannot effectively solve this type of problem.   For this new problem of sequential treatments, the causal effect estimation task can be transformed into a graph learning task based on a heterogeneous graph and directed acyclic graph. First, it constructs a biased heterogeneous graph with self-supervised learning, including many disconnected sub-graphs. Each sub-graph represents one unit and all its potential paths. Second, the learned heterogeneous graph is a typical directed acyclic graph, an architecture that processes information according to the flow defined by the partial order. Based on the practical implications of this DAG, bidirectional processing is utilized. A path may be processed to estimate the outcome at the end of the path by the natural order, and another is used to reconstruct the original feature by the reversed order.

\subsubsection{4.1.2 Structured Treatments}

In many practical situations, treatments are naturally structured, such as medical prescriptions (text), protein structures (graph), and computed tomography scans (image). Traditional treatment effect estimation methodologies typically use separate prediction heads for each treatment option so that the influence of the treatment indicator variable might be lost in the high-dimensional network representations. Extending this idea directly to structured treatments would not only be computationally expensive but would also not be able to make use of treatment features or learn treatment representations \cite{kaddour2021causal}.

GraphITE \cite{harada2021graphite} learns representations of graph treatments for CATE estimation. They propose to utilize graph neural networks while mitigating observation biases using Hilbert-Schmidt Independence Criterion regularization, which increases the independence of the representations of the targets and treatments. Inspired by the Robinson decomposition, which has enabled flexible CATE estimation for binary treatments, \cite{kaddour2021causal} propose the Generalized Robinson Decomposition (GRD) from which they extract a pseudo-outcome that targets the causal effect. A generalization of the GRD to treatments that can be vectorized as a continuous embedding. This GRD reveals a learnable pseudo-outcome target that isolates the causal component of the observed signal by eliminating confounding associations.

In addition, there is a growing methodological literature investigating how images should be integrated to estimate treatment effect \cite{pawlowski2020deep,castro2020causality} in the observational data. An image-based treatment effect model is proposed by using a deep probabilistic modeling framework \cite{jerzak2022image}. They develop a method that estimates latent clusters of images by identifying images with similar treatment effects distributions. The model also emphasizes an image sensitivity factor that quantifies the importance of image segments in contributing to the mean effect cluster prediction, obtained via Monte Carlo using the approximate posterior distribution over the clustering.

\subsection{Covariate}

In the treatment effect estimation task, the selection bias is the greatest challenge, which is the phenomenon that the distribution of the observed group is not representative of the group we are interested in. Confounder variables affect units' treatment choices, which leads to selection bias. This phenomenon exacerbates the difficulty of counterfactual outcome estimation, as we need to estimate the control outcome of units in the treated group based on the observed control group, and to estimate the treated outcome of units in the control group based on the observed treated group. The procedure for handling the selection bias is called covariate adjustment \cite{yao2021survey}.

As more covariates are collected in observational data, we face different types of covariates, such as confounders (observed and hidden), adjustment, instrumental, and spurious variables. However, more data do not mean better estimation results. Therefore, in this section, we discuss this topic from three aspects: (1) feature selection; (2) feature representation disentanglement; (3) hidden confounders.

\subsubsection{4.2.1 Feature Selection}

A common approach for covariate adjustment is using the propensity score, i.e., the probability of a unit being assigned to a particular level of treatment, given the background covariates \cite{rosenbaum1983central}. In covariate adjustment, although including all confounders is essential, this does not mean that including more variables is always better~\cite{chu2020matching,greenland2008invited,schisterman2009overadjustment}. For example, conditioning on \textit{instrumental} variables that are associated with the treatment assignment but not with the outcome except through treatment can increase both bias and variance of estimated causal effects~\cite{myers2011effects_instrumental}. Conditioning on \textit{adjustment} variables that are predictive of outcomes but not associated with treatment assignment is unnecessary to remove bias while reducing variance in estimated causal effects~\cite{sauer2013review}. Therefore, the inclusion of instrumental variable can inflate standard errors without improving bias, while the inclusion of adjustment variable can improve precision \cite{shortreed2017outcome,wilson2014confounder,lin2015regularization,zigler2014uncertainty}.

\cite{kuang2017treatment} propose a Data-Driven Variable Decomposition (D2VD) algorithm, which can automatically separate confounders and adjustment variables with a data-driven approach where a regularized integrated regression model is presented to enable confounder separation and ATE estimation simultaneously. \cite{chu2023estimating} propose a deep adaptive variable selection-based propensity score method (DAVSPS) by using representation learning and adaptive group LASSO. The key idea of DAFSPS is to combine the data-driven learning capability of representation learning and variable selection consistency of adaptive group LASSO to improve the estimation of the propensity score by selecting confounders and adjustment variables while removing instrumental and spurious variables. The framework of DAVSPS contains two major steps: outcome prediction with group LASSO and propensity score estimation with adaptive group LASSO. Step One uses a deep neural network (DNN) with group LASSO to predict the outcome and obtain the initial weight estimates for each covariate. Step Two uses a DNN classification model to estimate propensity scores with adaptive group LASSO, under which the weighted penalty is based on initial weight estimates obtained from step one. Therefore, DAVSPS can automatically select covariates predictive of the outcome (i.e., confounder and adjustment variables) while removing covariates independent of the outcome (i.e., instrumental and spurious variables) in propensity score estimation.

\begin{figure}
  \centering
  \includegraphics[width=1\columnwidth]{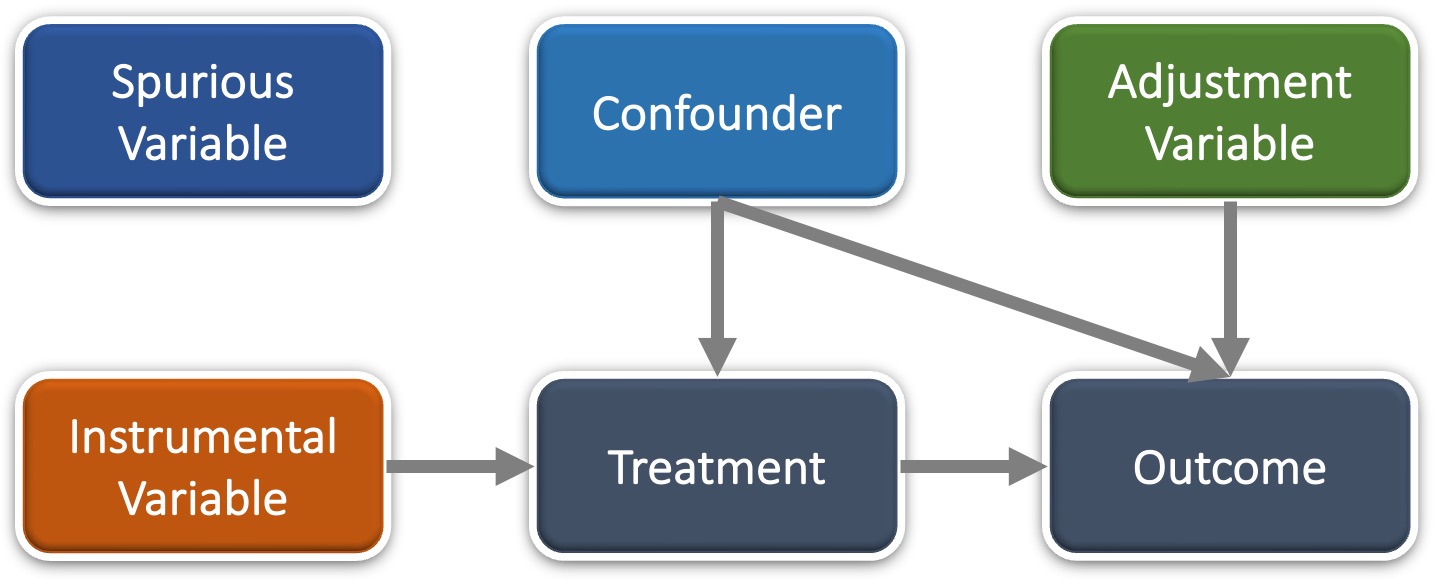}
  \caption{\label{fig:relationship} The relationships among treatment, confounder, outcome, instrumental, adjustment, and spurious variables.
  }
\end{figure}

\subsubsection{4.2.2 Feature Representation Disentanglement}

For a simple feature representation disentanglement, i.e., confounders and non-confounders, \cite{wu2020learning} propose a synergistic learning framework to identify confounders by learning decomposed representations of both confounders and non-confounders, and balance confounders with sample re-weighting technique, and simultaneously. Then, a more detailed disentangled representation learning method \cite{hassanpour2020learning} decomposes covariates into three latent factors, including instrumental $\Gamma$, confounding $\Delta$, and adjustment $\Upsilon$ factors. They assume that the random variable $X$ follows an unknown joint probability distribution $Pr( X | \Gamma, \Delta, \Upsilon )$, treatment $T$ follows $Pr( T | \Gamma, \Delta )$, and outcome $Y$  follows $Pr ( Y | \Delta, \Upsilon )$, where $\Gamma$, $\Delta$, and $\Upsilon$ represent the three underlying factors that generate an observational dataset. Correspondingly, the selection bias is induced by factors $\Gamma$ and $\Delta$, where $\Delta$ represents the confounding factors between $T$ and $Y$. \cite{zhang2021treatment} propose a variational inference approach to simultaneously infer latent factors from the observed variables, disentangle the factors into three disjoint sets corresponding to the instrumental, confounding, and adjustment factors, and use the disentangled factors for treatment effect estimation. However, it remains an open problem how to learn the underlying disentangled factors precisely. Specifically, previous methods may fail to obtain independent disentangled factors, which is necessary for identifying treatment effects. \cite{cheng2022learning} propose Disentangled Representations for Counterfactual Regression via Mutual Information Minimization (MIM-DRCFR), which uses a multi-task learning framework to share information when learning the latent factors and incorporates MI minimization learning criteria to ensure the independence of these factors.

\subsubsection{4.2.3 Hidden Confounders} Due to the fact that identifying all of the confounders is impossible in practice, the strong ignorability assumption is usually untenable. If a confounder is hidden or unmeasured, it is impossible in the general case without further assumptions to estimate the treatment effect on the outcome \cite{pearl2009causality}. By leveraging big data, it becomes possible to find a proxy for the hidden or unmeasured confounders by exploring the relationship between the hidden confounders, their proxies, the treatment, and the outcome. For example, Causal Effect Variational Autoencoder (CEVAE) \cite{louizos2017causal} is based on Variational Autoencoders (VAE), which follows the causal structure of inference with proxies. It can simultaneously estimate the unknown latent space summarizing the confounders and the causal effect. 

In addition, recent studies have shown that the auxiliary network information among data can be utilized to mitigate the confounding bias. The network information, which serves as an efficient structured representation of non-regular data, is ubiquitous in the real world. Advanced by the powerful representation capabilities of various graph neural networks, networked data has recently received increasing attention~\cite{kipf2016semi,velivckovic2017graph,velickovic2019deep,jiang2019censnet}. Therefore, it can also be used to help recognize the patterns of hidden confounders. A network deconfounder~\cite{guo2019learning} is proposed to recognize hidden confounders by combining the graph convolutional networks~\cite{kipf2016semi} and counterfactual regression~\cite{shalit2017estimating}. Unlike networked data in traditional graph learning tasks, such as node classification and link prediction, the networked data under the causal inference problem has its particularity, i.e., imbalanced network structure. \cite{chu2021graph} propose a Graph Infomax Adversarial Learning (GIAL) model for treatment effect estimation, which makes full use of the network structure to capture more information by recognizing the imbalance in network structure. However, the above works assume that the observational data and the relations among them are static, while in reality, both of them will continuously evolve over time, i.e., time-evolving networked observational data. \cite{ma2021deconfounding} propose a novel causal inference framework Dynamic Networked Observational Data Deconfounder (DNDC), which learns dynamic representations of hidden confounders over time by mapping the current observational data and historical information into the same representation space.

\subsection{Outcome}

The foremost challenge to treatment effect estimation with observational data is to handle the imbalance in the covariates with respect to different treatment options, which is caused by selection bias. Recent causal effect estimation methods~\cite{johansson2016learning,shalit2017estimating,li2017matching} have built a strong connection with domain adaptation by enforcing domain invariance with distributional distances such as the Wasserstein distance and maximum mean discrepancy. Inspired by metric learning, some methods~\cite{yao2018representation} use hard samples to learn representations that preserve local similarity information and balance the data distributions. In~\cite{zhang2020learning}, the authors argue that distribution invariance is often too strict a requirement and, they propose to use counterfactual variance to measure the domain overlap. Motivated by information theory, \cite{chu2022learning} propose to learn the Infomax and Domain-Independent Representations. It utilizes the mutual information between the global feature representations and individual feature representations, and the mutual information between feature representations and treatment assignment predictions in order to maximally capture the common predictive information for both treatment and control groups. 

For these domain adaptation methodologies based on the potential outcome framework (POF), the model aims to learn the \emph{domain-invariant representations} i.e., transformations of features, such that the treatment and control groups are approximately indistinguishable in the representation space~\cite{ben2007analysis}. Despite the popularity of domain adaptation for POF, the sufficient support assumption \cite{ben2012hardness} for domain adaptation uncovers intrinsic limitations of learning invariant representations when it comes to the shift in support of domains \cite{liu2021cycle}. The positivity assumption is an essential assumption in causal effect estimation, and it supports the strong sufficient support assumption for domain adaptation \cite{zhao2019learning,johansson2019support}. However, the positivity assumption is by no means guaranteed to hold in practice due to the following two reasons. First, high-dimensional data often contain redundant or irrelevant information for predicting the outcome but still helps to distinguish the treatment and control groups. Second, variables distributed differently across treatment groups are usually critical for prediction. 

Besides, for the domain adaptation problem under POF settings, seeking the optimal metric to measure the distance between the treatment and control groups remains unsettled. The choice of distance metrics is highly dependent on the characteristics of data distributions and the hyperparameters of regularization terms for imbalance mitigation. Especially, even with the same selection bias, there is no consensus among different metrics in terms of balancing data distributions~\cite{zhang2020learning}.

Finally, we argue that regularizing representations to be domain-invariant is too strict, in particular when domains (e.g., treatment and control groups) are partially overlapped~\cite{zhang2020learning}. Several studies show that the empirical risk minimization only on factual data outperforms domain-invariant representation learning algorithms. Therefore, enforcing domain-invariant can easily remove predictive information and lead to a loss in predictive power, regardless of which type of domain divergence metrics is employed~\cite{alaa2018limits}. These observations motivate us to relax the positivity assumption and develop a new and unified paradigm for treatment effect estimation, such that we could avoid the choice dilemma of domain divergence metrics and also overcome the loss of predictive information. This is a promising and urgent direction for the treatment effect estimation task.

\section{Conclusion}
Causal inference is a developing field of academic research and various industrial applications. Recently, the blooming development of machine learning has brought new vitality into the causal inference area, not only the excellent progress on original problems but also the new research potentials and directions. In this survey, we comprehensively review emerging advances, challenges, and opportunities for the treatment effect estimation task from the three core components, i.e., treatment, covariates, and outcome.

\bibliographystyle{named}
\bibliography{ijcai23}

\end{document}